\documentclass[sigconf]{acmart}
\AtBeginDocument{%
  }

\setcopyright{acmlicensed}
\copyrightyear{2018}
\acmYear{2018}
\acmDOI{XXXXXXX.XXXXXXX}

\usepackage{amsmath,amssymb}
\usepackage{algorithmic}
\usepackage{graphicx}
\usepackage{textcomp}
\usepackage{xcolor}
\usepackage{url}
\usepackage{amsthm}
\usepackage{subfigure}
\usepackage{tabularx}
\usepackage{enumitem}
\usepackage{multicol}
\usepackage{multirow}
\usepackage{algorithm}
\usepackage{algorithmic}

\setcopyright{none}
\begin{document}
\title{Causal Graph Profiling via Structural Divergence for Robust Anomaly Detection in Cyber-Physical Systems} 

\author{Arun Vignesh Malarkkan}
\email{arun.malarkkan@asu.edu}
\affiliation{%
  \institution{Arizona State University}
  \city{Tempe}
  \state{Arizona}
  \country{USA}
}

\author{Haoyue Bai}
\email{haoyueba@asu.edu}
\affiliation{%
  \institution{Arizona State University}
  \city{Tempe}
  \state{Arizona}
  \country{USA}
}

\author{Dongjie Wang}
\email{wangdongjie@ku.edu}
\affiliation{%
  \institution{University of Kansas}
  \city{Lawrence}
  \state{Kansas}
  \country{USA}
}

\author{Yanjie Fu$^{{\dagger}}$}
\email{yanjie.fu@asu.edu}
\affiliation{%
  \institution{Arizona State University}
  \city{Tempe}
  \state{Arizona}
  \country{USA}
}

\thanks{\small $^{\dagger}$Corresponding author}

\begin{abstract}
With the growing complexity of cyberattacks targeting critical infrastructures such as water treatment networks, there is a pressing need for robust anomaly detection strategies that account for both system vulnerabilities and evolving attack patterns. Traditional methods—statistical, density-based, and graph-based models struggle with distribution shifts and class imbalance in multivariate time series, often leading to high false positive rates. 
To address these challenges, we propose CGAD: a Causal Graph-based Anomaly Detection framework designed for reliable cyberattack detection in public infrastructure systems. CGAD follows a two-phase supervised framework: causal profiling and anomaly scoring. 
First, it learns causal invariant graph structures representing the system's behavior under "Normal" and "Attack" states using Dynamic Bayesian Networks. 
Second, it employs structural divergence to detect anomalies via causal graph comparison by evaluating topological deviations in causal graphs over time. 
By leveraging causal structures, CGAD achieves superior adaptability and accuracy in non-stationary and imbalanced time series environments compared to conventional machine learning approaches. 
By uncovering causal structures beneath volatile sensor data, our framework not only detects cyberattacks with markedly higher precision but also redefines robustness in anomaly detection, proving resilience where traditional models falter under imbalance and drift. 
Our framework achieves substantial gains in F1 and ROC-AUC scores over best-performing baselines across four industrial datasets, demonstrating robust detection of delayed and structurally complex anomalies.
\end{abstract}
\maketitle
\section{Introduction}
Critical public infrastructures—including transportation systems, energy grids, water treatment facilities, healthcare services, and communication networks—form the backbone of societal functionality, public safety, and economic stability.
These infrastructures are increasingly vulnerable to cyberattacks, which can lead to cascading service disruptions, severe economic damage, and threats to public health and safety.
For example, in 2021, a cyberattack on the Oldsmar, Florida water treatment plant attempted to manipulate chemical levels, exposing critical vulnerabilities and prompting widespread, costly security upgrades across the sector~\cite{cervini2022don}.
Robust cyberattack detection in such infrastructures is essential to mitigate immediate threats and preserve system integrity. Yet, this task is particularly challenging due to the high-dimensional, temporally dependent nature of sensor data and severe class imbalance, where malicious behavior is rare compared to normal operations. These challenges are magnified in cyber-physical domains like industrial control systems and water treatment networks, where distributed sensors, dynamic operational contexts, and evolving attack surfaces complicate reliable anomaly detection.
In this paper, we focus on \textit{Water Treatment Networks (WTNs)}—a critical class of industrial cyber-physical infrastructure that manages water purification and distribution through interconnected sensors, actuators, and communication modules. WTNs generate \textit{multivariate, high-dimensional time-series data} that reflect complex dependencies among physical and cyber components, making them a realistic and challenging testbed for cyberattack detection.
We formalize the task of cyberattack detection in WTNs as the identification and classification of anomalous segments in multivariate time-series data collected from distributed sensors, using historical attack labels as supervision signals. The goal is to develop models that can detect both known and unseen cyberattacks with high precision and robustness.

Addressing cyberattack detection within the Water Treatment Networks critically introduces two seminal research challenges:

\noindent\textbf{C1: Modeling invariant anomaly influence structures:} How can we learn a robust representation of inter-sensor influence structures that remain invariant to attack strategies and class imbalance?

\noindent\textbf{C2: Fast anomaly scoring:} Given the learned structure, how can we efficiently quantify the degree of anomalous deviation in incoming time segments with training or supervision?

In defending WTNs against cyberattacks, prior literature includes both research-driven and applied anomaly detection approaches, each with specific limitations.
\textbf{Applied techniques} focus on efficiency and simplicity: 
\textit{1) Knowledge-based detection} uses expert-defined thresholds (e.g., chlorine levels, valve pressure), but lacks adaptability and scalability~\cite{sommer2010outside}. 
\textit{2) Statistics-based detection} employs probabilistic or distance-based thresholds, yet struggles with nonlinear dependencies and manual tuning~\cite{ahmed2016survey,montgomery2007introduction}. 
\textit{3) Unsupervised detection} methods such as clustering and anomaly scoring (e.g., k-means, Isolation Forests) scale well but suffer from high false positives due to limited context~\cite{chandola2009anomaly, breunig2000lof}. 
\textit{Supervised detection} methods like SVMs and random forests require labeled data and generalize poorly to novel attacks~\cite{chandola2009anomaly}.
\textbf{Research-oriented methods} emphasize modeling complexity and structural reasoning: 
\textit{4) Deep learning and graph-based detection} (e.g., LSTMs, GNNs) offer expressive temporal and relational modeling~\cite{malhotra2015long, 10.1145/3196494.3196546, 10415739, Boniol_2020}, but suffer from high computational cost, limited interpretability, and reliance on heuristic graph construction. 
\textit{5) Causal graph fusion approaches} (e.g., SMV-CGAD) combine dense and sparse views for improved robustness~\cite{10.1145/3627673.3680096, malarkkan2025rethinking, malarkkan2025incremental}, but require domain priors and use deep graph classifiers that reduce transparency.

\noindent\textbf{Our Insights: a causal graph-enabled profiling-scoring perspective.}
While prior work has explored correlation- or signal-based graph structures, few approaches model the causal mechanisms underlying WTN operations. We argue that \textbf{causal graphs}, represented as \textbf{Directed Acyclic Graphs (DAGs)}, offer a principled way to capture stable normal and anomaly dependency structures that remain invariant across shifting operational conditions and evolving attacks. This angle enhances robustness to distribution shifts while offering actionable insights by elucidating the propagation of anomalies across system components.

\begin{figure}[h]
\centering
\includegraphics[width=\linewidth]{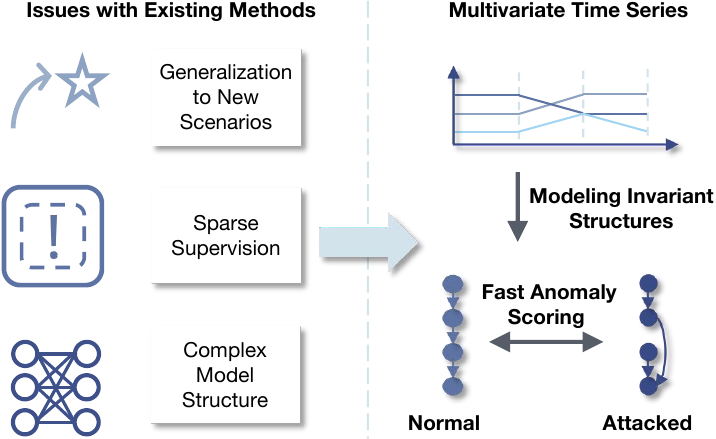}
\vspace{-0.5cm}
\caption{Leveraging causal structural differences between normal and attack conditions in WTNs for superior anomaly detection compared to existing traditional methods.}
\end{figure}

\noindent\textbf{Summary of Proposed Method: CGAD – A Causal Profiling-Scoring Framework.}
To this end, we propose \textbf{CGAD}, a causal graph-based framework for anomaly detection from multivariate sensor time series. CGAD has two main components:
1) \textbf{Causal Anomaly Profiling:} We learn two causal DAGs to represent the normal and anomalous system states using observational time-series data.
2) \textbf{Causal Deviation Scoring:} We develop a DAG-to-DAG distance metric to compute an anomaly score for test data segments. Specifically, if the causal structure of a test segment closely resembles the anomaly DAG, it is flagged as anomalous; otherwise, it is considered normal. Our DAG-DAG distance metric accounts for both structural topology (e.g., edge presence/absence) and causal strength (e.g., edge weights), providing a holistic and efficient way to measure deviation between causal models.
This framework enables fast, accurate, and robust detection of cyberattacks and is designed to be robust to class imbalance and distribution shifts commonly observed in real-world datasets.

\noindent\textbf{Our Contributions.}
We address the pressing problem of cyberattack detection in Water Treatment Networks using multivariate sensor time series. Our main contributions are:

\begin{itemize}
    \item We introduce \textbf{CGAD}, a novel causal graph-based framework that explicitly models and compares normal and abnormal causal structures in WTN data.
    \item We formulate a \textbf{two-phase profiling-scoring pipeline} that first learns DAGs to represent system states and then computes a \textbf{causal divergence score} via DAG-DAG comparison.
    \item We demonstrate that CGAD provides \textbf{robust and accurate detection} of cyberattacks in WTNs, significantly improving performance over existing baselines.
\end{itemize}

\section{Problem Statement}

\noindent
\textbf{Water Treatment Network (WTN):}  
Consider a water treatment network \(W\) instrumented with \(K\) sensors monitoring various treatment process stages~\cite{9338292}.  
Continuous sensor streams are partitioned into \(N\) non-overlapping intervals, each with \(M\) time-aligned measurements, yielding a sensor stream data sequence \(X = [X_1, \dots, X_N]\), where \(X_i \in \mathbb{R}^{M \times K}\). 
Each segment \(X_i\) is labeled by \(y_i \in \{0,1\}\), with \(y_i = 1\) denoting an attack and \(y_i = 0\) as normal operation.

\noindent
\textbf{The Detection Task:}  
Let the multivariate time series data be segmented into \(N\) non-overlapping intervals, each denoted by \(X_i \in \mathbb{R}^{M \times K}\), where \(M\) is the number of time steps per segment and \(K\) is the number of sensors. The dataset is given by \(D = \{(X_1, y_1), \dots, (X_N, y_N)\}\), where \(y_i \in \{0,1\}\) indicates whether segment \(X_i\) corresponds to an \textit{Attack} (1) or \textit{Normal} (0) system state.  
The AI task is to learn a model that can detect cyberattacks in WTN from multivariate time series. The model must account for both immediate disruptions and delayed attack effects, ensuring robust detection across short-term and delayed structural deviations.

\noindent
\textbf{Objective:}  
 We propose a robust, efficient framework for segment-level cyberattack detection via \textit{structural causal learning}. By modeling inter-sensor causal links, our method detects deviations from normal causal dynamics, enhancing generalization to novel attacks.

\begin{figure*}[h!]
    \centering
    \includegraphics[width=\textwidth]{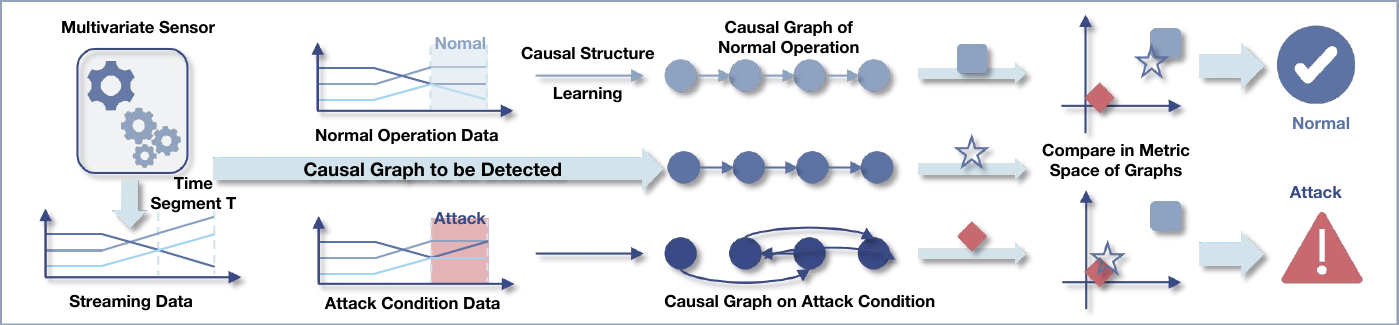}
    \vspace{-0.5cm}
    \caption{Overview of the CGAD Framework for Cyberattack Detection.  
    Phase 1 - Causal Graph Learning: Causal Profiling learns causal graphs for "Normal" and "Attack" states using DYNOTEARS algorithm.  
    Phase 2 - Anomaly Detection via Graph Comparison: Anomaly Scoring segments data, infers a graph for each segment, and classifies segments based on Structural Hamming Distance (SHD) to reference graphs.
    }
\end{figure*}

\section{Proposed Method}

\subsection{Framework Overview}

Figure 2 illustrates the architecture of our proposed CGAD framework, which comprises two core phases:

\noindent\textbf{(P1) Causal Profiling:} We learn invariant causal structures from multivariate sensor time series by estimating Dynamic Bayesian Networks (DBNs) representing the system under distinct operational states. Specifically, two causal graphs are constructed: one for \textit{Normal} operation and one for \textit{Attack} conditions. This profiling captures stable, structurally insightful inter-sensor dependencies that reflect underlying system dynamics.

\noindent\textbf{(P2) Anomaly Scoring:} Streaming data is segmented temporally, and for each segment, a causal graph is inferred using the same procedure. We then compute the Structural Hamming Distance (SHD) between the segment’s graph and each of the two reference graphs. The segment is classified as \textit{Attack} if its causal structure is closer (lower SHD) to the \textit{Attack} causal graph, and \textit{Normal} otherwise.

This two-phase causal profiling and scoring pipeline leverages robust causal representations to detect anomalies efficiently and with increased transparency. Unlike traditional correlation-based methods, CGAD explicitly models causal mechanisms, enabling resilient detection across diverse attack patterns and operational shifts. Although only a single attack graph is used as a ground-truth, it captures recurring structural disruptions that are common across different attacks, allowing CGAD to generalize effectively to unseen threats through structural graph comparisons.

\vspace{-0.2cm}
\subsection{Causal Profiling - Causal DAG Learning for Time-Series Data}
In this phase, we uncover the causal structure of multivariate sensor data to distinguish genuine anomalies from spurious correlations. Accurate modeling of these structures is critical, as cyberattacks in Water Treatment Networks (WTNs) often manifest through temporally extended and structurally coherent disruptions.
To capture both instantaneous and time-lagged dependencies, we model the system as a Dynamic Bayesian Network (DBN), represented as a time-aware Directed Acyclic Graph (DAG). In this DAG, each \textbf{node} corresponds to a specific sensor reading (e.g., valve pressure, flow rate, chemical concentration) at a given time step. \textbf{Edges} represent directed causal influence: an edge from node \(A_t\) to node \(B_{t+1}\) implies that the value of sensor \(A\) at time \(t\) has a causal effect on sensor \(B\) at the next time step \(t+1\). Intra-slice edges model instantaneous dependencies within a segment, while inter-slice edges capture delayed or sequential effects across time. Together, the DAG encodes the underlying operational rules and inter-sensor interactions governing system behavior.
We employ the \textbf{DYNOTEARS} algorithm~\cite{pamfil2020dynotears}, a state-of-the-art approach for learning DBNs from time-series data. Unlike static DAG learning methods such as NOTEARS~\cite{zheng2018dags}, DYNOTEARS explicitly incorporates temporal structure and can effectively model delayed, persistent, and cascading effects that characterize real-world cyberattacks~\cite{9338292}. Using this approach, we construct two causal graphs: one representing the normal operating state (\(G_{\text{Normal}}\)) and the other capturing the structural footprint of anomalous behavior (\(G_{\text{Attack}}\)). These graphs serve as causal profiles for the scoring phase, enabling structure-aware detection of deviations in unseen data segments.

Let \(\mathcal{X} = [X_1, X_2, \ldots, X_n]\) denote the multivariate time-series data, segmented into \(n\) non-overlapping time windows. Each segment \(X_i \in \mathbb{R}^{M \times K}\) contains \(M\) time steps across \(K\) sensors in a WTN.
We design the temporal layout such that anomalous segments are bounded by normal segments both before and after the attack interval. Formally, we define the sequence as:
\[
X_i = [N_1, N_2, A_3, A_4, \ldots, A_{m-2}, N_{m-1}, N_m],
\]
where \(N_j\) denotes a segment in a \textit{Normal} state, and \(A_j\) denotes a segment under an \textit{Attack} state.

This structured segmentation allows us to construct two ground-truth causal graphs: \(G_{\text{Normal}}\) and \(G_{\text{Attack}}\), which characterize inter-sensor dependencies under normal and attack conditions, respectively. To learn these graphs, we apply the \textbf{DYNOTEARS} algorithm~\cite{pamfil2020dynotears}, an extension of NOTEARS~\cite{zheng2018dags}, designed to uncover both inter-slice (temporal lag) and intra-slice (instantaneous) causal dependencies from time-series data.

Let \(X \in \mathbb{R}^{T \times K}\) denote the design matrix formed by concatenating the sensor readings from all segments within a class (either normal or attack), and let \(Y \in \mathbb{R}^{T \times K}\) denote the corresponding time-lagged matrix. The Structural Equation Model (SEM) is defined as:
\begin{equation}
    X = XW + YA + Z,
\end{equation}
where \(W \in \mathbb{R}^{K \times K}\) is the intra-slice (instantaneous) adjacency matrix, \(A \in \mathbb{R}^{K \times K}\) is the inter-slice (temporal) adjacency matrix, and \(Z\) is a residual noise matrix.

The goal is to estimate sparse, interpretable matrices \(W\) and \(A\) such that \(W\) defines an acyclic structure. This constrained optimization objective is achieved by:
\begin{equation}
    \min_{W, A} \, f(W, A) \quad \text{s.t.} \quad h(W) = 0,
\end{equation}
with:
\begin{equation}
    f(W, A) = \ell(W, A) + \lambda_W \|W\|_1 + \lambda_A \|A\|_1,
\end{equation}
where \(\ell(W, A)\) denotes the squared loss term, and \(\lambda_W, \lambda_A\) are hyperparameters controlling the sparsity via \(\ell_1\)-norm regularization.
To enforce acyclicity of the intra-slice graph, DYNOTEARS uses the continuous constraint:
\begin{equation}
    h(W) = \operatorname{tr}(e^{W \circ W}) - d = 0,
\end{equation}
where \(\circ\) denotes the Hadamard (element-wise) product and \(d = K\) is the number of nodes in the graph.

At the end of this phase, we obtain the two causal graphs—\(G_{\text{Normal}}\) and \(G_{\text{Attack}}\)—which serve as reference structures for anomaly detection via graph comparison in the subsequent scoring phase. During training, we learn two reference graphs, \(G_{\text{Normal}}\) and \(G_{\text{Attack}}\), by applying DYNOTEARS to concatenated normal and attack segments, respectively. At inference time, we apply DYNOTEARS independently to each unseen segment \(X_t\) to construct a corresponding causal graph \(G_t\), which is then compared to the reference graphs using structural divergence for anomaly scoring.

\subsection{Anomaly Scoring via Structural Divergence}

\noindent\textbf{Causal Graph Construction:}  
In this phase, we perform causal profiling by learning a causal graph for each time segment of the test data and comparing it to reference causal graphs representing normal and attack states. This enables time-segmented anomaly detection, addressing limitations of traditional point-level detection methods. Specifically, point-level approaches are prone to false positives in imbalanced settings, overly sensitive to benign signal fluctuations, and often fail to detect temporally delayed or system-wide attack effects. In contrast, our causal graph-based strategy captures context-aware structural dependencies, allowing for more robust and interpretable anomaly detection. For each test segment \(T\), we apply the DYNOTEARS algorithm to infer the segment-specific causal graph \(G_T\), representing both instantaneous and lagged inter-sensor relationships within that segment.

\noindent\textbf{Structural Comparison:}  
Given the learned causal graph \(G_T\) for a test segment, we assess its similarity to the reference graphs \(G_{\text{Normal}}\) and \(G_{\text{Attack}}\) using a principled graph-theoretic metric: the Structural Hamming Distance (SHD). SHD quantifies the dissimilarity between two DAGs as the minimum number of edge additions, deletions, or direction reversals between them. This allows us to robustly measure deviations in causal structure without relying on distributional or feature-based matching.

Formally, given two graphs \(G_1 = (V, E_1)\) and \(G_2 = (V, E_2)\) with the same set of nodes \(V\), the SHD is defined as:
\begin{equation}
\text{SHD}(G_1, G_2) = |E_1 \setminus E_2| + |E_2 \setminus E_1|,
\end{equation}
where \(E_1 \setminus E_2\) denotes the set of directed edges present in \(G_1\) but not in \(G_2\), and vice versa.

For the test graph \(G_T\) for time segment \(T\), we compute:
\begin{equation}
SHD_{TA} = \text{SHD}(G_T, G_{\text{Attack}}), \quad SHD_{TN} = \text{SHD}(G_T, G_{\text{Normal}})
\end{equation}
representing the graph distance to the attack and normal references.

\noindent\textbf{Decision Rule:}  
We assign the predicted class label \(\hat{y}_T\) by determining which reference graph the test graph more closely resembles:

\begin{equation}
\hat{y}_T =
\begin{cases}
1, & \text{if } SHD_{TA} < SHD_{TN}, \\
0, & \text{otherwise,}
\end{cases}
\end{equation}
where \(\hat{y}_T = 1\) denotes an \textit{Attack} state and \(\hat{y}_T = 0\), a \textit{Normal} state.

In essence, if the causal structure \(G_T\) of the test segment is more similar to the attack reference graph \(G_{\text{Attack}}\) than to the normal graph \(G_{\text{Normal}}\), we classify it as anomalous. Otherwise, it is deemed normal. This approach enables CGAD to generalize to previously unseen attack types, as many cyberattacks introduce recurring causal disruptions, even if their surface signal patterns differ.

\noindent This framework effectively maps each test segment to an operational state using global structural characteristics rather than raw features. Notably, \textbf{SHD focuses on structural rather than distributional differences, making CGAD resilient to transient noise, nonstationarity, and high class imbalance, three defining challenges in cyber-physical anomaly detection.}

\section{Experiments}
We conduct experiments to answer the following questions:

\noindent\textbf{RQ1}:  Does our method improve anomaly detection compared with baseline methods?   

\noindent\textbf{RQ2}: What are the impacts of causal graph learning and graph divergence metrics in our framework?  Is structural hamming distance the most effective graph comparison metric for anomaly detection?

\noindent\textbf{RQ3}: Is our method robust over different data conditions? 

\noindent\textbf{RQ4}: How sensitive is our method to key hyperparameters?

\noindent\textbf{RQ5}:  What is the computational cost of our method?

\subsection{Experimental Setup}
\subsubsection{Datasets.} 
We use four real-world datasets: 
\textbf{1) SWaT} \cite{7469060}: A 11-day water treatment plant testbed data collected from 51 interconnected sensors. The dataset contains 16 fault events over the period, which had 7 days of "Normal" system status and 4 days of "Attack" system status. 
\textbf{2) WADI} \cite{10.1145/3055366.3055375}: Water Distribution testbed dataset collected over 16 days from 123 actuators/sensors. 15 attack events were recorded in the last 2 days of the collection period with "Attack" system status. 
\textbf{3) Tennessee Eastman (TE)} \cite{4519-z502-19}: A chemical process simulation dataset with 52 sensors and 20 anomalies. Training spans 25 hours, testing 48 hours, with measurements every three minutes. Table 1 shows dataset statistics, train-test split, and causal graph node count. 
\textbf{4) Server Machine Dataset (SMD) \cite{10.1145/3292500.3330672}} is a server monitoring dataset over 5 weeks monitoring 28 server machines from 38 sensors. It is one of the largest public datasets for anomaly detection in multivariate time-series data.

\begin{table}[h!]
\vspace{-5pt}
\renewcommand\arraystretch{1.0}
	\centering
	\caption{DATASET STATISTICS.}
        \vspace{-0.3cm}
        \scalebox{0.70}{
	\begin{tabular}{c|c|c|c|c|c}
		\hline
		  Dataset & Status & \# Features & \# Normal Data & \# Attack Data & Normal to Attack Ratio        \\ \hline
            \textbf{SWaT} & Normal & 51 & 495000 & 0 & 0        \\
                 &Attack & 51 & 395298 & 54621 & 7:1        \\
            \hline
            \textbf{WADI} & Normal & 123 & 1048571 & 0 & 0        \\
                 &Attack & 123 & 162824 & 9977 & 16:1        \\
            \hline
            \textbf{TE} & Normal & 52 & 450000 & 0 & 0        \\
                 &Attack & 52 & 222500 & 22800 & 10:1        \\
            \hline
            \textbf{SMD} & Normal & 38 & 708405 & 0 & 0        \\
                 &Attack & 38 & 678950 & 29470 & 25:1        \\
            \hline
	\end{tabular}
        }
	\label{tab:dataset statistics}
\end{table}
\vspace{-3pt}
\subsubsection{Evaluation Metrics}
\textbf{Point-adjusted F1-Score \cite{febrinanto2023entropy}: } In multivariate time-series data, attack events often form continuous segments, making pointwise anomaly detection less effective \cite{9525836}. Our method requires a time-interval segment to capture the underlying causal structure. We use the point-adjusted F1-score, where a segment is labeled anomalous if any point within it is an anomaly, and the F1-score is computed based on the performance of our method on the entire segment.
\textbf{ROC-AUC:} AUC, the area under the Receiver Operating Characteristic (ROC) curve, provides a measure of a model's ability to distinguish positive from negative samples.
\textbf{PRC-AUC:} PRC-AUC focuses on the performance of the model on the anomalous class. This is particularly useful when the dataset is imbalanced, like in anomaly detection.

\subsubsection{\textit{Baseline Algorithms}}
We compare our method with the following baseline algorithms:
\textbf{1) One-Class Support Vector Machine (OC-SVM)}~\cite{yin2014fault} – 
Supervised anomaly detection algorithm employing a kernel-based hyperplane decision boundary for classification of anomaly samples.
\textbf{2) Isolation Forest}~\cite{xu2017improved} – 
Ensemble supervised anomaly detection isolating anomalies via subsampling on streaming data.
\textbf{3) Deep Support Vector Data Description (Deep SVDD)} \cite{ZHANG20211} – 
Deep learning anomaly detection via hypersphere learning.
\textbf{4) Hybrid CNN-LSTM}~\cite{10296546} – 
An efficient unsupervised anomaly detection framework. 
\textbf{5) Spatio-Temporal Outlier Detection (STOD)}~\cite{9338292} – 
A spatio-temporal outlier detection. 
\textbf{6) Angle Based Outlier Detection (ABOD)}~\cite{10.1145/1401890.1401946} – 
Outlier detection using angle metrics.
\textbf{7) Empirical Cumulative Distributed Functions for Outlier Detection (ECOD)} \cite{9737003} – 
A parameter-free, interpretable unsupervised anomaly detection method. 
\textbf{8) Lightweight On-Line Detector of Anomalies (LODA)} \cite{pevny2016loda} – 
Efficient unsupervised ensemble of weak detectors.
\textbf{9) SMV-CGAD}~\cite{10.1145/3627673.3680096}-
Spectral multi-view causal graph anomaly detection with dense/sparse graph structures and deep graph convolution.

\subsubsection*{\textit{Implementation Details}}

\noindent
\textbf{Assumptions.} We assume anomalous events are rare relative to normal behavior, introducing class imbalance that impairs conventional detection methods.
\noindent
\textbf{Data Setup.} Time-series data are split into non-overlapping 15-minute segments. Models are trained on historical data and tested on future segments to preserve causal validity.
\noindent
\textbf{Causal Learning.} We use DYNOTEARS from \texttt{CausalNex}~\cite{Beaumont_CausalNex_2021} to learn Dynamic Bayesian Networks, with time-lags: 4 (SWaT), 3 (WADI), 4 (TE), 1 (SMD). Gaussian noise is added to attack data to regularize graph learning.
\noindent
\textbf{Baselines.} Competing methods use \texttt{PyOD}~\cite{zhao2019pyod} with default settings.
\noindent
\textbf{Hardware.} Experiments were conducted on Intel i9-12900HK CPU, 32\,GB RAM, and NVIDIA RTX 4090 GPU.
\noindent
The code repository is available in \href{CGAD-repo}{https://anonymous.4open.science/r/CGAD-4E18/}

\begin{table*}
\renewcommand\arraystretch{1.02}
\centering
\caption{Overall Performance Across All Datasets}
\vspace{-0.3cm}
\label{tab:performance_analysis}
\small{
\begin{tabularx}{\textwidth}{c|X|X|X|X|X|X}
\hline
\multirow{2}{*}{Methods} & \multicolumn{3}{c|}{\textbf{SWAT}} & \multicolumn{3}{c}{\textbf{WADI}} \\
\cline{2-7}
                         & $F1_{PA}$ & ROC-AUC & PRC-AUC & $F1_{PA}$ & ROC-AUC & PRC-AUC  \\
\hline
STOD                     & 0.6955 & 0.8420 & 0.7013   & 0.4126 & 0.5176 & 0.5660  \\
Deep-SVDD                & 0.1729 & 0.6618 & 0.2927   & 0.1591 & 0.4938 & 0.2798  \\
CNN-LSTM                 & 0.6129 & 0.6881 & 0.5997   & 0.7891 & 0.8138 & 0.7798  \\
ECOD                     & 0.2403 & 0.7819 & 0.3320   & 0.4077 & 0.6897 & 0.5137   \\
LODA                     & 0.6942 & 0.8972 & 0.7154   & 0.2251 & 0.5516 & 0.2920   \\
ABOD                     & 0.1194 & 0.5000 & 0.2321   & 0.1126 & 0.5000 & 0.05   \\
One-class SVM            & 0.7385 & 0.6632 & 0.7441   & 0.5343 & 0.4208 & 0.6121  \\
Isolation Forest         & 0.7412 & 0.8426 & 0.8122   & 0.6434 & 0.6487 & 0.6850   \\
SMV-CGAD                 & 0.7532 & 0.8211 & 0.7355   & 0.6679 & 0.7831 & 0.7077   \\
\hline
CGAD - DAGNOTEARS        & 0.1156 & 0.5427 & 0.2322   & 0.5879 & 0.7120 & 0.5470   \\
\textbf{CGAD-DYNOTEARS}  & \textbf{0.7807} & \textbf{0.8611} & \textbf{0.7388}   & \textbf{0.8913} & \textbf{0.9015} & \textbf{0.8504}   \\
\hline
\multicolumn{7}{c}{} \\[-1.5ex] 
\hline
\multirow{2}{*}{Methods} & \multicolumn{3}{c|}{\textbf{TE}} & \multicolumn{3}{c}{\textbf{SMD}} \\
\cline{2-7}
                         & $F1_{PA}$ & ROC-AUC & PRC-AUC  & $F1_{PA}$ & ROC-AUC & PRC-AUC  \\
\hline
STOD                     & 0.6421 & 0.7280 & 0.7106  & 0.6779 & 0.8076 & 0.6840  \\
Deep-SVDD                & 0.4801 & 0.5679 & 0.4680  & 0.3911 & 0.5448 & 0.4794   \\
CNN-LSTM                 & 0.6778 & 0.8328 & 0.6990  & 0.6490 & 0.5030 & 0.5720   \\
ECOD                     & 0.4003 & 0.5000 & 0.4222  & 0.5827 & 0.5100 & 0.6100   \\
LODA                     & 0.4228 & 0.5007 & 0.5102  & 0.2656 & 0.5030 & 0.2866   \\
ABOD                     & 0.1827 & 0.4730 & 0.2560  & 0.6244 & 0.5027 & 0.5009   \\
One-class SVM            & 0.8125 & 0.8231 & 0.8441  & 0.8443 & 0.8288 & 0.8624   \\
Isolation Forest         & 0.7421 & 0.8102 & 0.7684  & 0.8501 & 0.8734 & 0.8681   \\
SMV-CGAD                 & 0.7389 & 0.8002 & 0.7556  & 0.8395 & 0.8030 & 0.8112  \\
\hline
CGAD - DAGNOTEARS        & 0.3512 & 0.5011 & 0.3323   & 0.4670 & 0.5802 & 0.5017   \\
\textbf{CGAD-DYNOTEARS}  & \textbf{0.8297} & \textbf{0.8516} & \textbf{0.7888}   & \textbf{0.8626} & \textbf{0.8542} & \textbf{0.9004}  \\
\hline
\end{tabularx}
}
\end{table*}

\subsection{Experimental Results}

\subsubsection{RQ1: Overall Performance}
To answer \textbf{RQ1} , Table \ref{tab:performance_analysis} shows our  method (\textbf{CGAD}) in overall outperforms baseline methods on the \textbf{SWaT}, \textbf{WADI}, \textbf{TE} and \textbf{SMD} datasets in terms of five metrics: $F1_{PA}$, ROC-AUC, and PRC-AUC. 
The experiments demonstrate four insights: 1) CGAD outperforms correlation-based and density-based methods by capturing stable cause-effect patterns that reflect true system dynamics. 2) Despite its lightweight design, CGAD surpasses deeper models like SMV-CGAD, highlighting the strength of structural divergence over fused deep representations. 3)  CGAD maintains high F1 and ROC-AUC across all datasets, demonstrating resilience to distribution shifts and data imbalance.  4)  CGAD offers structurally derived alerts based on causal deviation, crucial for actionable insights in high-stakes infrastructure systems.
While CGAD-DYNOTEARS shows strong overall performance, its effectiveness relies on the assumption that causal structures can be reliably and robustly estimated from segmented time-series data. In highly nonstationary settings where causal discovery algorithms fail to recover meaningful structures, CGAD's performance may degrade. But SMV-CGAD, by integrating dense and sparse views with deep representations, can be more robust in such cases by capturing implicit patterns even when explicit structures are unreliable.

\subsubsection{RQ2: Study of Causal Graph Learning and Graph Divergence Metrics} To answer \textbf{RQ2}, we develop an ablation study to examine two technical components. 

\noindent\underline{\it Effect of Causal Graph Learning (Phase 1):} 
Our method considers temporal causal structures from multivariate time series. The baseline method, DAGs with NO TEARS~\cite{zheng2018dags}, performs poorly due to ignoring temporal dependencies, latent confounders, and delayed attack effects in cyber-physical systems. In contrast, DYNOTEARS, which models regressive temporal dynamics, significantly improves AUC and $F1_{PA}$, confirming the importance of temporal modeling for robust anomaly detection."

\noindent\underline{\it Effect of Graph Divergence Metrics (Phase 2):} We detect anomalies by measuring structural change between causal graphs using a graph comparison metric.  We compare several alternatives, including \textbf{Jaccard similarity} (edge set overlap), and \textbf{Laplacian spectral distance}.  We observe that while these metrics yield comparable trends in anomaly localization, they fail to capture subtle structural deviations. 
However, Structural Hamming Distance balances accuracy with efficiency and achieves the best trade-off between detection fidelity and runtime cost. 

\subsubsection{RQ3: Robustness Check}
\begin{figure*}[htbp]
\centering
\subfigure[SWaT: F1 Score Distribution]{
    \includegraphics[width=4.3cm,height=3.5cm]{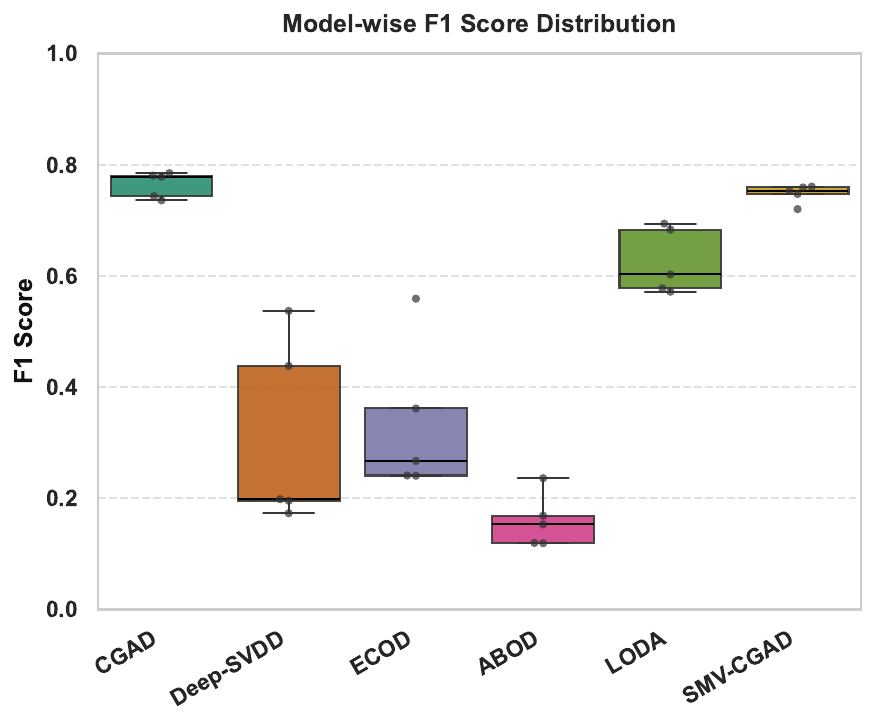}
}
\hspace{-3mm}
\subfigure[SWaT: ROC-AUC Distribution]{
    \includegraphics[width=4.3cm,height=3.5cm]{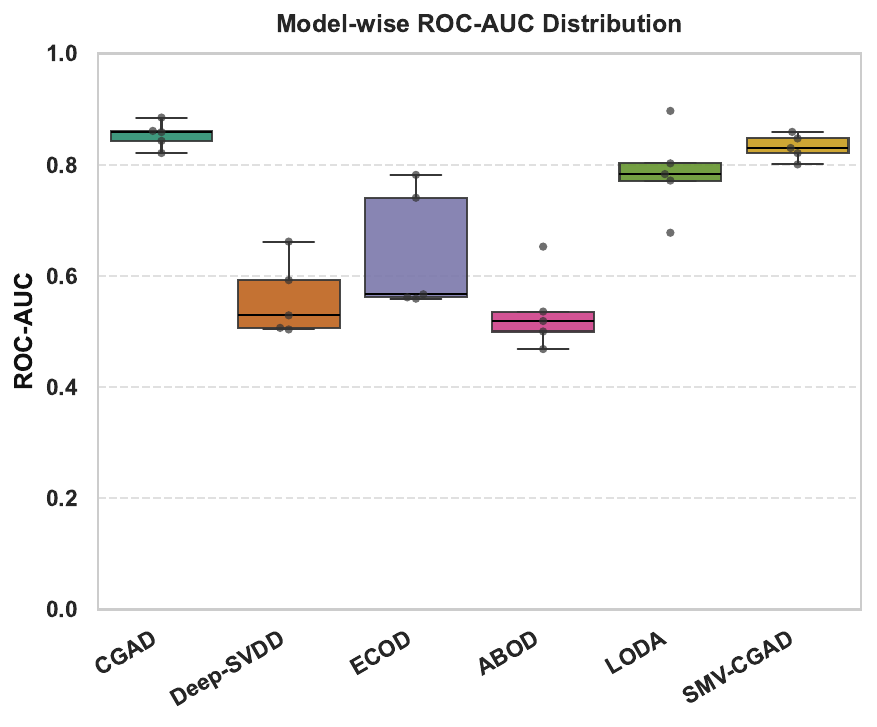}
}
\hspace{-3mm}
\subfigure[WADI: F1 Score Distribution]{
    \includegraphics[width=4.3cm,height=3.5cm]{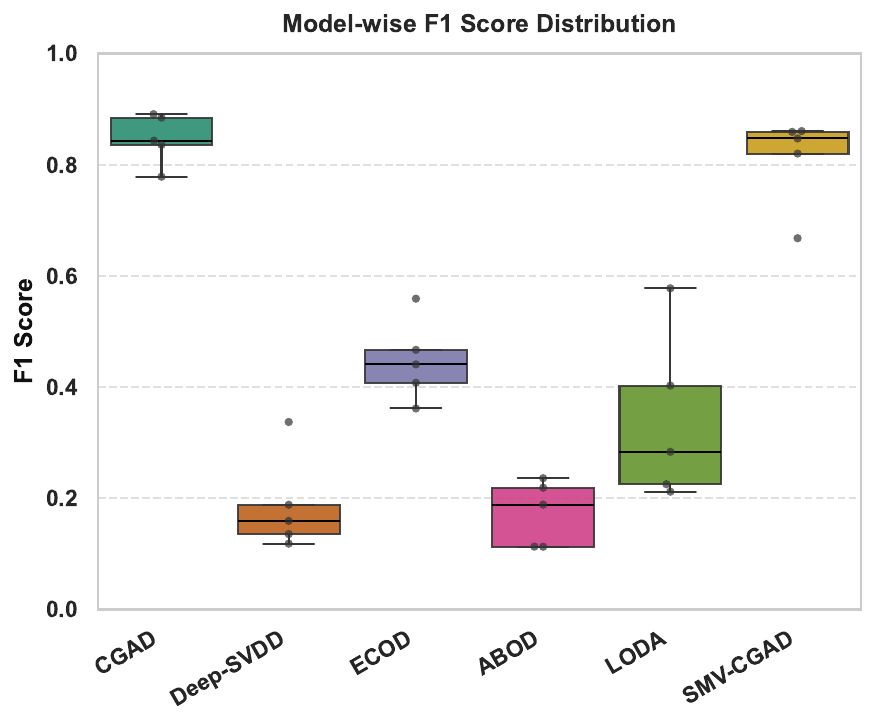}
}
\hspace{-3mm}
\subfigure[WADI: ROC-AUC Distribution]{
    \includegraphics[width=4.3cm,height=3.5cm]{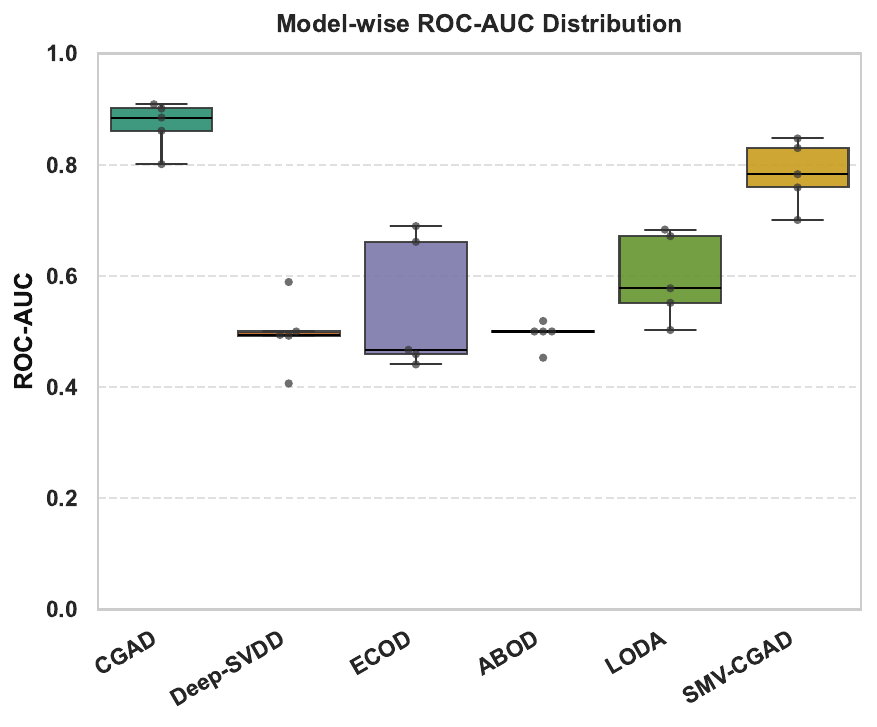}
}
\vspace{-0.2cm}
\caption{
\textbf{Model robustness analysis across test subsets on the SWaT and WADI datasets.} 
Each boxplot illustrates the distribution of detection performance across five temporal subsets for each model.
}
\label{fig:robustness_sw_wadi}
\vspace{-0.2cm}
\end{figure*}

To answer RQ3, we evaluate the robustness of CGAD by assessing whether CGAD can consistently detect anomalies via a causal perspective across multiple subsets of temporal streams on the SWaT and WADI datasets.
Figure \ref{fig:robustness_sw_wadi} shows that CGAD achieves stable performance across multiple balanced data subsets, indicating the robustness of causal graphs learned from different samples of the same underlying distribution. However, a notable drop in all evaluation metrics is observed when a data subset is significantly imbalanced. This degradation underscores a known limitation of the DYNOTEARS algorithm—its sensitivity to the availability of high-quality, causally relevant data during structure learning.
Additionally, we observe that the omission of Gaussian noise during the causal graph construction for the 'Attack' state further reduces performance. This suggests that introducing controlled noise into the attack data aids in capturing the stochastic nature of attack-induced perturbations, thereby improving the generalization capacity of the learned causal representations. These findings highlight the importance of data quality, balance, and controlled regularization for robust causal discovery in adversarial cyber-physical environments.

\subsubsection{RQ4: Impact of Time-Lag Parameter in DYNOTEARS}
\begin{figure}[htbp]
\vspace{-0.1cm}
\centering
\subfigure{
\includegraphics[width=4cm,height=3cm]{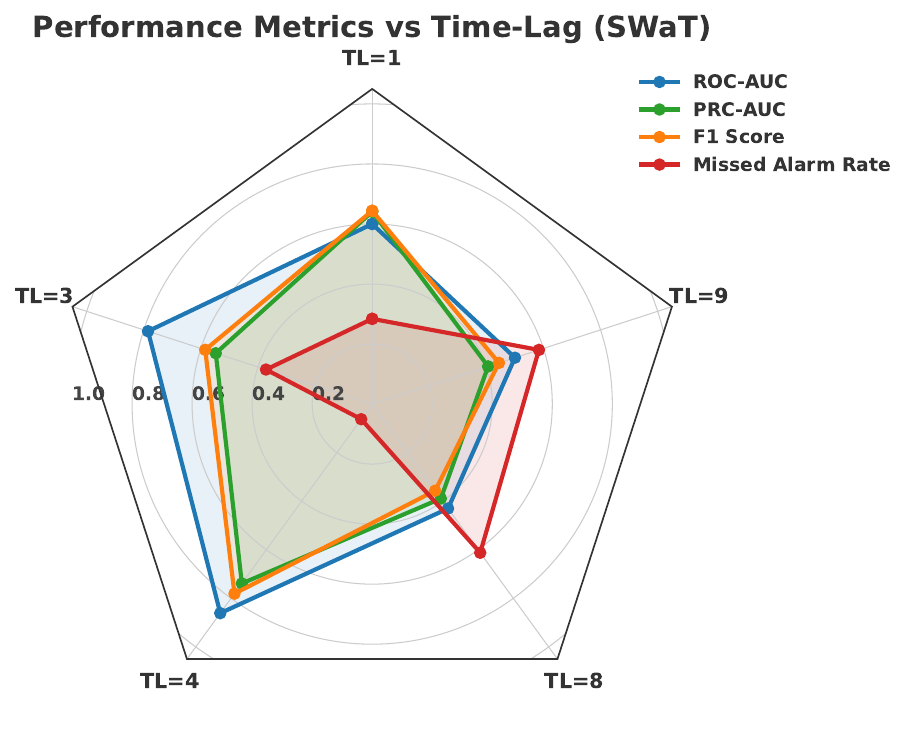}
}
\hspace{-3mm}
\subfigure{ 
\includegraphics[width=4cm,height=3cm]{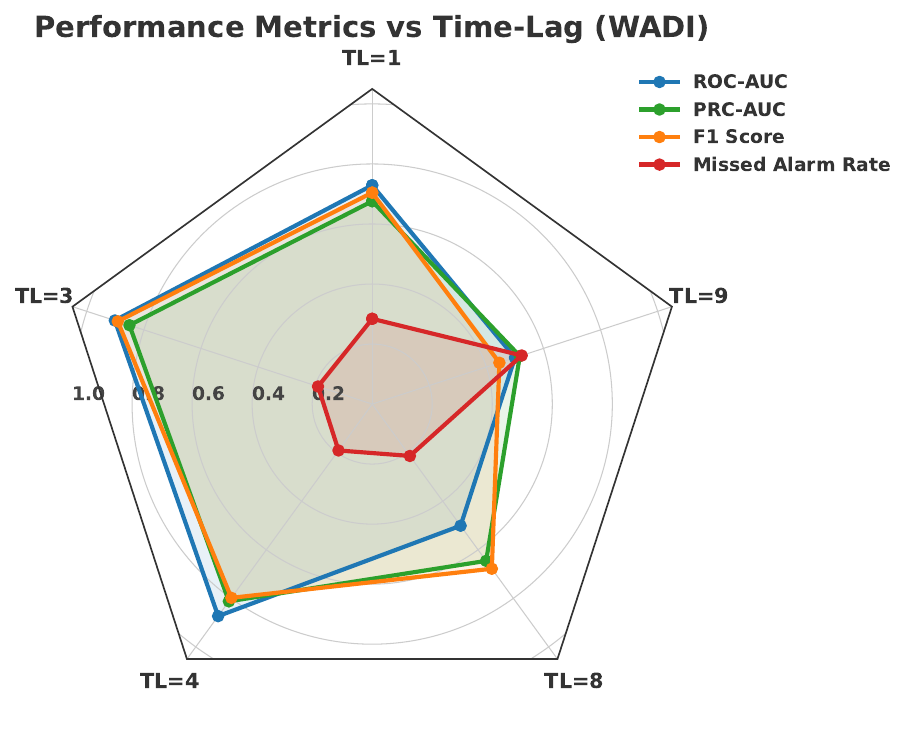}
}
\hspace{-3mm}
\subfigure{
\includegraphics[width=4cm,height=3cm]{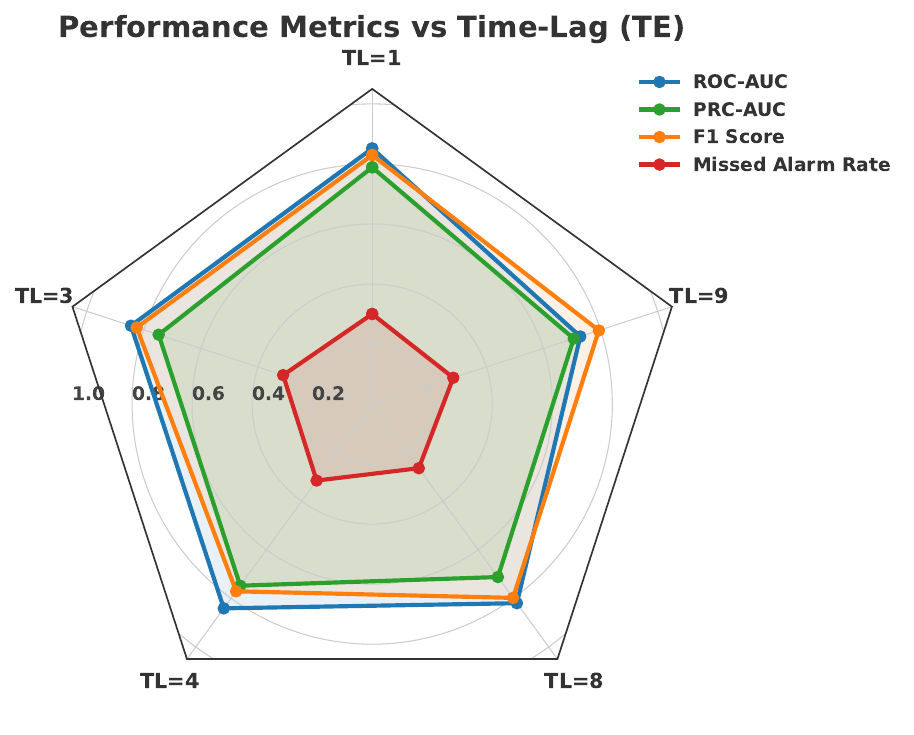}
}
\hspace{-3mm}
\subfigure{ 
\includegraphics[width=4cm,height=3cm]{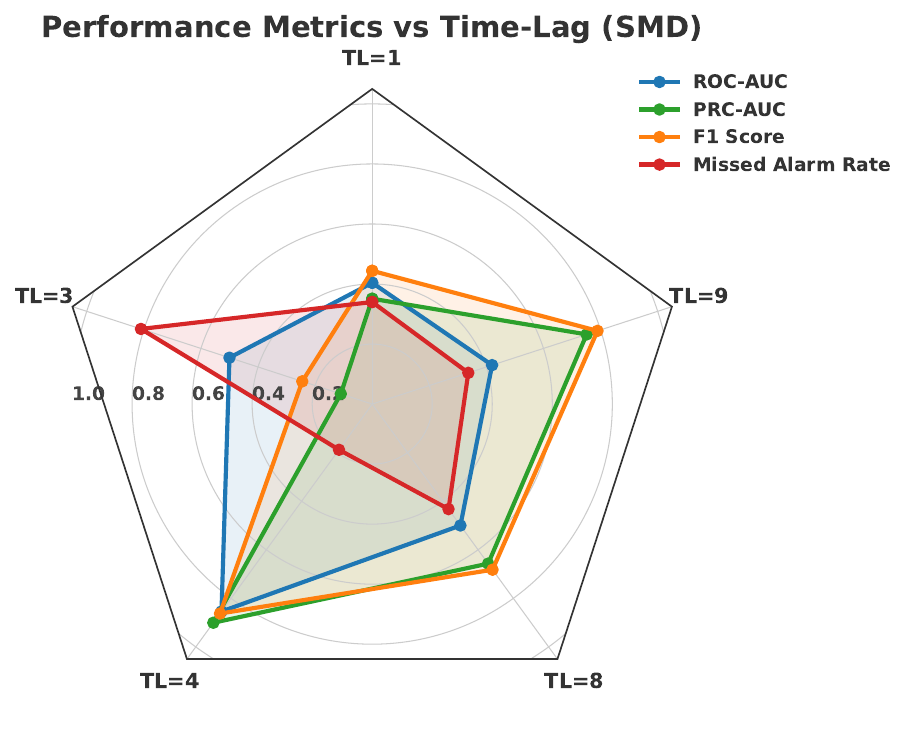}
}
\vspace{-0.5cm}
\caption{CGAD Performance vs. Time-lag parameter.}
\label{fig:timelag}
\end{figure}

The time-lag parameter in the DYNOTEARS algorithm is critical in modeling the temporal dependencies that characterize delayed effects of cyberattacks. It defines the maximum historical window considered when estimating temporal causal relationships.
To answer RQ4, we conduct a sensitivity analysis by manually tuning the time-lag parameter across multiple values and evaluating the detection performance. Figure \ref{fig:timelag} shows that optimal performance is achieved with a time-lag of 4, 3, 4, and 1 for the SWaT, WADI, TE, and SMD datasets, respectively. Smaller lag values tend to miss delayed causal effects, while larger values risk overfitting and introducing instability in the learned structures.
This analysis confirms that the effectiveness of DYNOTEARS in CGAD is highly dependent on appropriate lag selection, which must be tailored to the specific temporal characteristics of the underlying system dynamics.

\subsubsection{RQ5: Runtime Analysis and Computational Efficiency}
\begin{figure}[htbp]
\centering
\subfigure[SWaT: Train-Test time comparison]{
    \includegraphics[width=4cm,height=3.2cm]{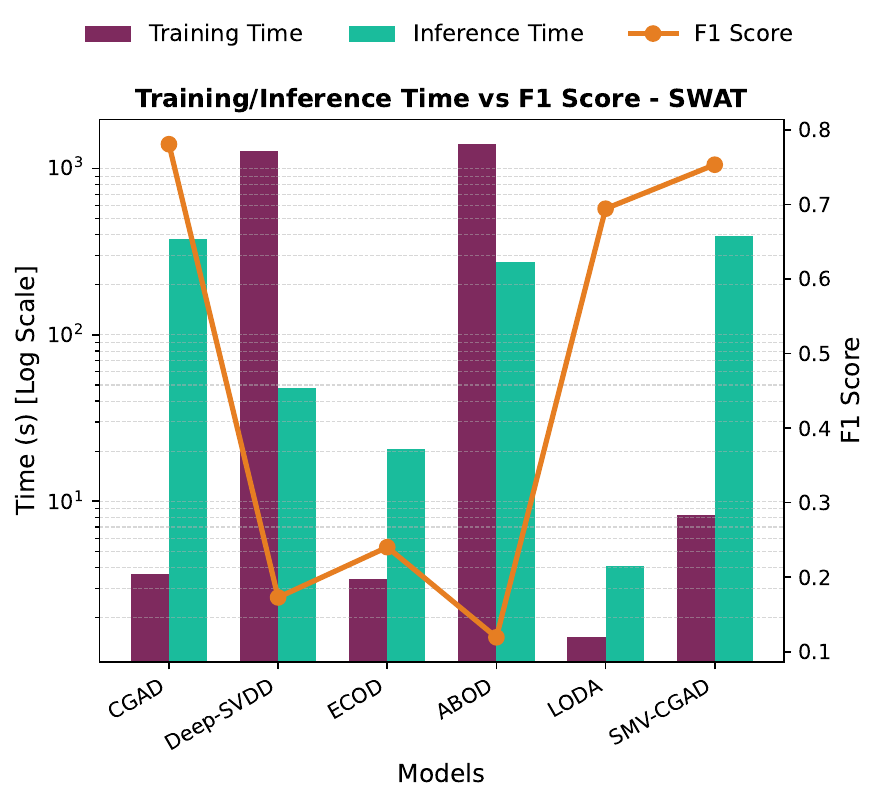}
}
\hspace{-3mm}
\subfigure[WADI: Train-Test time comparison]{
    \includegraphics[width=4cm,height=3.2cm]{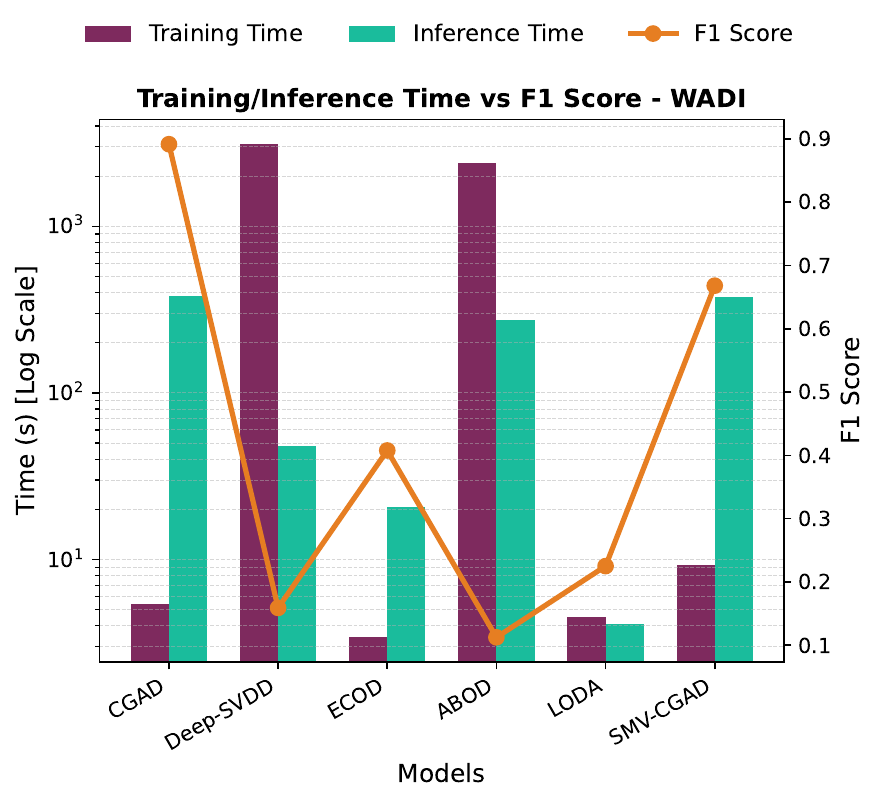}
}

\vspace{-0.5cm}
\caption{Runtime analysis for training and inference phases.}
\label{fig:runtime_logscale}
\vspace{-0.2cm}
\end{figure}

To answer RQ5, we analyzed CGAD's computational efficiency by comparing its training and inference times against all baselines on the SWaT and WADI datasets. Figure \ref{fig:runtime_logscale} demonstrates CGAD's competitive training time, comparable to LODA and significantly faster than Deep-SVDD and ABOD. This efficiency stems from DYNOTEARS' score-based optimization, which avoids exhaustive graph search and scales effectively with variables.
Inference time, which involves learning causal graphs for each test segment, is moderately higher due to the segment-wise causal discovery. However, the trade-off is justified by the model’s superior transparency and accuracy, especially in the presence of complex or delayed attack patterns. Overall, CGAD offers a practical balance between computational efficiency and detection robustness, making it well-suited for deployment in real-time industrial monitoring environments.

\subsubsection{Discussions: Scalability Considerations and Future Directions}
Causal graph learning, by design, seeks to uncover underlying cause-effect relationships among features—in this case, the sensors within cyber-physical systems. A key advantage of the CGAD framework is that as the data scale increases, the resulting causal representations become more expressive, capturing richer and more generalized system behaviors. This scalability potential distinguishes CGAD from traditional methods that struggle to model such complexity effectively.
Our runtime analysis demonstrates that CGAD achieves relatively low training time compared to several baseline models. While test-time inference is moderately slower due to the need to learn causal structures per time segment, the DYNOTEARS algorithm’s score-based optimization remains tractable even for high-dimensional datasets. This computational efficiency makes CGAD a viable option for deployment in real-world settings that demand real-time monitoring over time-series data.
Nonetheless, several scalability challenges remain, consistent with broader limitations of machine learning in industrial settings:

\begin{itemize}
    \item \textbf{Computational Cost:} Causal graph learning over large datasets can be computationally intensive. To mitigate this, we propose continual causal learning, selective modeling on representative subsets, domain-informed constraints, and early stopping to reduce overhead at scale.
    
    \item \textbf{Data Quality:} Noisy, inconsistent, or biased sensor data may compromise causal discovery. Robust preprocessing, outlier handling, and noise-tolerant graph learning are essential for accurate structure estimation.

    \item \textbf{Operational Constraints:} Deployment may be hindered by regulatory and organizational barriers. These require interpretable models and collaboration with domain stakeholders.

\end{itemize}

To address these concerns, we are actively pursuing industry collaborations to validate CGAD in pilot deployments. As a forward-looking extension, we aim to develop a \textit{continuous causal learning} module that incrementally updates the causal graph in response to real-time sensor feedback. Such a mechanism would enhance adaptability and resilience to concept drift, ensuring sustained performance in dynamic operational environments.

\section{Related Work}
Our framework aligns at the intersection of \textit{causal discovery in time-series}, \textit{graph-based reasoning}, and \textit{robust anomaly detection}.

\noindent\textbf{Causal Discovery in Time-Series.}
Learning temporal causality in CPS requires uncovering both lagged and instantaneous dependencies, often obscured by noise. Early VAR-based Granger causality methods lack scalability and fail under nonlinearity. Recent advances like TCDF~\cite{nauta2019causal} provide neural or score-based causal discovery, modeling multivariate dynamics with time lags. Gong et al.~\cite{gong2023causal} survey modern methods (e.g., PCMCI, attention-based causal transformers~\cite{zhang2023attention}) that achieve state-of-the-art results. However, many still rely on strong stationarity assumptions.

\noindent\textbf{Graph-Based Reasoning in CPS.}
Graph abstractions represent complex infrastructure states in CPS. Gorawski et al.~\cite{8970674} model smart city infrastructure as sensor networks, while Kirchheim et al.~\cite{kirchheim2023towards} show structured representations improve anomaly explanation. Deep graph models (e.g., GCFormer~\cite{xing2023gcformer}, GDN~\cite{deng2021graph}) capture spatio-temporal dependencies but rely on predefined topologies, limiting transparency. Recent work explores causal interventions to improve cross-graph generalization~\cite{10.1145/3696410.3714712}, highlighting the value of structural invariance in anomaly-prone environments.

\noindent\textbf{Anomaly Detection in Cyber-Physical Systems.}
Anomaly detection in CPS is challenging due to scarce labels and high-dimensional, noisy streams. Traditional statistical or density-based methods~\cite{chandola2009anomaly} suffer from high false positives. Deep models like Deep SVDD~\cite{ruff2018deep} or Anomaly Transformer~\cite{xu2022anomaly} improve recall but struggle with explainability and drift. Semi-supervised frameworks~\cite{VILLAPEREZ2021106878} reduce supervision needs but often lack robustness to evolving attack patterns. CGAD addresses these by modeling causal invariants across system states, enabling reliable detection with actionable insights.

\section{Conclusion Remarks}
In this work, we addressed the critical challenge of detecting cyberattacks in Water Treatment Networks by proposing a novel causal graph-based anomaly detection framework, \textbf{CGAD}. It operates in two phases: (i) \textit{Causal Profiling}, employing DYNOTEARS to learn ground-truth causal graphs for "Normal" and "Attack" system behaviors; and (ii) \textit{Causal Scoring}, where segmented sensor data's causal graphs are inferred and compared to references via structural divergence.
Through extensive experimentation on four real-world cyber-physical datasets, we demonstrate CGAD achieves high detection accuracy and robustness to class imbalance, distributional shifts, and delayed attack manifestations. By leveraging causal stability over correlation, CGAD offers explainable, efficient, and generalizable anomaly detection in complex time-series environments. While promising, the framework presents scalability challenges, particularly in causal graph learning for high-dimensional or noisy data. Addressing these requires efficient graph learning strategies and adaptation for evolving system behaviors.
Looking ahead, we envision extending CGAD for real-time deployment in large-scale industrial and critical infrastructure systems. Domains like Finance, Healthcare, and Cybersecurity can greatly benefit from CGAD’s explainable causal reasoning and structural anomaly detection capabilities. Future work will focus on continuous causal learning and collaborative testing with industry partners to ensure robustness and adaptability in dynamic operational settings.

\section{GenAI Usage Disclosure}
Generative AI tools were used to refine sections of this paper for improved clarity, coherence, and grammatical accuracy. All core ideas, algorithms, experiments, and analyses were conceived, implemented, and validated solely by the authors.


\bibliographystyle{ACM-Reference-Format}
\bibliography{main}

\end{document}